\documentclass[sigconf]{acmart}
\usepackage{xcolor}
\usepackage{caption}
\usepackage{subcaption}

\AtBeginDocument{%
  \providecommand\BibTeX{{%
    \normalfont B\kern-0.5em{\scshape i\kern-0.25em b}\kern-0.8em\TeX}}}

\setcopyright{rightsretained}
\copyrightyear{2024}
\acmYear{2024}
\acmDOI{}

\acmConference[HRI '24: X-HRI Workshop]{Explainability for Human-Robot Collaboration (X-HRI ’24)}{March 11, 2024}{Boulder, CO, USA}

\acmBooktitle{HRI '24: Explainability for Human-Robot Collaboration (X-HRI) March 11, 2024 Boulder, CO, USA} 
\acmISBN{}
\begin{document}


\title{Leveraging Counterfactual Paths for Contrastive Explanations of POMDP Policies}

\author{Benjamin Kraske}
\authornote{Both authors contributed equally to this research.}
\email{Benjamin.Kraske@colorado.edu}
\orcid{0009-0000-9618-1392}
\affiliation{%
  \institution{University of Colorado Boulder}
  \streetaddress{3375 Discovery Drive}
  \city{Boulder}
  \state{CO}
  \country{USA}
  \postcode{80305}
}

\author{Zakariya Laouar}
\authornotemark[1]
\email{Zakariya.Laouar@colorado.edu}
\orcid{0009-0006-7942-7323}
\affiliation{%
  \institution{University of Colorado Boulder}
  \streetaddress{3375 Discovery Drive}
  \city{Boulder}
  \state{CO}
  \country{USA}
  \postcode{80305}
}

\author{Zachary Sunberg}
\email{Zachary.Sunberg@colorado.edu }
\orcid{0000-0001-9707-3035}
\affiliation{%
  \institution{University of Colorado Boulder}
  \streetaddress{3375 Discovery Drive}
  \city{Boulder}
  \state{CO}
  \country{USA}
  \postcode{80305}}

\renewcommand{\shortauthors}{Kraske*, Laouar*, \& Sunberg}

\begin{abstract}
    As humans come to rely on autonomous systems more, ensuring the transparency of such systems is important to their continued adoption. Explainable Artificial Intelligence (XAI) aims to reduce confusion and foster trust in systems by providing explanations of agent behavior. Partially observable Markov decision processes (POMDPs) provide a flexible framework capable of reasoning over transition and state uncertainty, while also being amenable to explanation. This work investigates the use of user-provided counterfactuals to generate contrastive explanations of POMDP policies. Feature expectations are used as a means of contrasting the performance of these policies. We demonstrate our approach in a Search and Rescue (SAR) setting. We analyze and discuss the associated challenges through two case studies.   
\end{abstract}



\received{6 March 2024}

\maketitle

\section{Introduction}

As artificial intelligence is increasingly adopted in settings that involve human supervision, it is increasingly important that end users are able to understand the reasoning behind the decisions made by such systems. This is especially important when artificial intelligence is used in mission-critical roles, such as search and rescue \cite{ray2023human}. The ability of an expert to ask questions and resolve confusion about the methods and results of such a system may make the difference between the adoption and appropriate trust of a system, and the mistrust and disuse of a system. Explainable Artificial Intelligence (XAI) seeks to enhance trust and enable transparency in these systems. Ideally, these autonomous systems should not only perform at a high level but also maintain sufficient transparency such that clear explanations of system behavior can be provided to end users. 

The partially observable Markov decision process (POMDP) provides a flexible framework for reasoning over state and transition uncertainty. POMDPs have been applied to problems ranging from air collision avoidance \cite{kochenderfer2012next} to cancer screening \cite{ayer2012or}. POMDPs are capable of capturing complex domains with millions of states \cite{ye2017despot} while accounting for uncertainty over these states. (PO)MDPs also lend themselves well to explanations, with more inherent transparency than black box methods. While much of XAI focuses on black box methods \cite{ras2022explainable}, explanations for model-based methods, specifically explainable planning, are increasingly a focus \cite{chakraborti_emerging}.

When interfacing autonomy with an end user, structured interaction can be useful to establish transparency in and trust of the system \cite{miller2019explanation}. Specifically, contrastive explanations such as \textit{"A performs better than B because C"} can be very useful in gaining the trust of end users \cite{chakraborti_emerging}. \textit{Counterfactuals}, or alternatives, (e.g., to executed actions or policies) provide insight into what could have happened under modified components of the system and can help make systems more interpretable \cite{byrne2019counterfactuals}. However, the application of counterfactuals to different classes of problems is not always immediately intuitive. In this work, we tackle this challenge and explore the use of user-given counterfactuals to provide contrastive explanations for POMDP policies. We demonstrate the approach in the context of a Search and Rescue (SAR) POMDP example and discuss associated challenges. 

 We first provide a brief overview of explainable planning. We then propose a methodology for counterfactual path explanations for a SAR POMDP domain, and conclude with illustrative examples and a following discussion.

\subsection{Explainable Planning} 
Planning explanations can be organized into model-based algorithm-agnostic and algorithm-specific explanations \cite{chakraborti_emerging}. Model-based explanations assume that an algorithm has solved for an optimal policy, and that any user confusion would be as a result of mistranslations of the user's preferences and not a result of algorithm limitations or misunderstanding of algorithm reasoning. However, explaining the characteristics of the policy can give more insight into the quantitative reasoning of the algorithm and enhance the interpretability of the system. Interpretability amounts to understanding the outcome of an algorithm in terms of the quantitative flow of information \cite{doran2017does}. Comprehensibility entails comprehending the outcome when explained using symbols such as environment landmarks and characteristics \cite{doran2017does}. 

Many works have sought to increase the explainability of planning, particularly in the context of MDPs. The objective of MDPs is to generate a policy that maximizes the (discounted) expected reward. Offering contrastive explanations in terms of the expected reward may be interpretable to users but may not enhance comprehensibility. Instead, it can be intuitive to evaluate a policy with respect to the expected feature occupancy, where features symbolize abstract components of the reward function \cite{elizalde2009generating, khan2009minimal}. This can enhance comprehensibility for the end user. \citet{luebbers2023autonomous} investigate the timing of contrastive justifications of paths generated by solving MPDs. \citet{soni2021not} utilize counterfactual queries to first build user profiles and then provide tailored explanations but do not provide contrastive explanations after inferring a given user.

POMDPs introduce an added layer of complexity by reasoning over uncertainty in the state. The true state is hidden and is only partially observable. \textit{Optimal} (i.e., with respect to reward) POMDP planning consists of branching on not only state transitions but also observations. This type of planning has the potential to seem unintuitive to an end user. As a result, many works have attempted to make POMDP reasoning more interpretable and comprehensible.

Several works explored interactions between the autonomy and the end user to enhance trust, yet do not provide explanations for agent behavior \cite{asmar2022collaborative, peltzer2023incorporating}.

Other works provide explanations in terms of the POMDP components. In \cite{wang2016impact, wang2018my}, the agent behavior is explained with respect to the POMDP's beliefs, rewards, transitions, and observations. However, the interaction between the autonomy and the user only occurs one way. The user is not able to provide counterfactuals to further understand the behavior. 

To assess counterfactuals for a POMDP (e.g. a user-defined open-loop path), contrastive explanations need to present expectations with respect to state transitions, observations, and initial belief. \citet{mazzi2023risk} analyze traces of a POMCP tree to reason about safety-critical belief-dependent decisions. This was conducted by representing undesired actions using rule templates and shields in the POMCP tree to avoid actions that violate belief thresholds.

We aim to provide a straightforward method of explaining policies by contrasting them against user-proposed counterfactuals via feature expectations. The following section discusses our methodology for generating these explanations for a SAR POMDP.

\section{Methodology}
\label{sec:meth}
The goal of this work is to provide an intuitive means for the user to better understand why a given path is executed in the SAR POMDP domain. Here, we focus on addressing differences between user and algorithm reasoning. Consider the following motivating example:

\begin{example} [Search and Rescue]
\label{example}
Consider a search and rescue scenario with a human rescuer and an autonomous search agent (e.g. a UAV) collaborating to find a missing person. To accomplish this, the human expresses an objective for the agent in terms of regions of interest and the agent plans policies according to this objective and the primary objective of locating the missing person. The agent is also constrained by a limited battery life which demands the agent to return to home before depleting. 

\end{example}

Our approach leverages the visual nature of this problem to acquire user feedback which informs explanations. We propose the following workflow for explanations of the POMDP policy given counterfactual user paths.

Given an executed POMDP policy, the user may have questions related to why the specific path was chosen, generally in contrast to another path. This provides an opportunity to explain the optimal POMDP policy in contrast to this alternative path. The user may express this counterfactual path by drawing it on a user interface or another means. This user path can then be translated into a sequence of actions, forming an open-loop policy, which is not dependent on observations or state transitions but is dependent on the horizon of the problem only. The performance of this policy is then compared to the optimal policy, forming the basis of an explanation. More specifically, an explanation describes why the optimal policy outperforms the user-proposed alternative and hence why the realized path differs from the user's expectations.

\subsection{Leveraging Feature Expectations}
A straightforward approach to policy explanation/justification is to convey that the optimal policy achieves the same or higher maximum expected reward than all other policies (that is, that the optimal policy is indeed optimal). In order to demonstrate that the policy is optimal, the expected reward of any proposed alternative policy could be compared against that of the optimal policy, demonstrating that the user could do no better than the optimal policy. However, while this style of explanation justifies the actions taken under the optimal policy, it does not provide any insight into why this policy accumulates the maximum expected reward. 

To enhance comprehensibility, we leverage features and weights to represent the components that contribute to the reward, giving insight into which problem objectives the algorithm satisfies. This is an extension of explainable planning literature, in which feature expectations and factored rewards are used to provide explanations for MDPs \cite{elizalde2009generating, khan2009minimal}.

\citet{choi_inverse_2011} present feature expectations for POMDPs, building on \citet{abbeel2004apprenticeship}. Let $\phi(s,a)$ be a feature occupancy function, where $s$ and $a$ are the state and action, respectively. This function returns a vector with entry $i$ equal to $1$ if a feature $i$ is occupied, and $0$ otherwise. Let $\alpha$ define a weighting for each feature such that the reward $R(s,a) = \alpha \cdot \phi(s,a)$. The feature expectation is defined as $\mu^\pi(b_0)=\mathbb{E}\left[\sum_{t=0}^{\infty} \gamma^t \phi\left(b_t, a_t\right) \mid \pi, b_0 \right]$, where $b$ is a belief distribution over states. Further, $\phi\left(b_t, a_t\right) = \sum_{s\in S}b(s_t)\phi(s,a_t)$, where $S$ is the state space of the POMDP and $b(s)$ is the probability of state $s$ in belief $b$. Then the value of a policy from some initial belief can be expressed as $V^\pi(b) = \alpha \cdot \mu(\pi)$ \cite{choi_inverse_2011}. In this way, feature expectations provide a means of separating the valuing of certain outcomes from the frequency of these outcomes. We leverage feature expectations for POMDPs as a means of explanations in light of features (which could be thought of as multiple objectives) that compose a reward function. For the SAR domain, these features include locations of interest, battery depletion, and locating a target.

This approach gives users insight into both the frequency of visited features and the value assigned to them, such that the user can better understand the contribution of each to the expected reward. Such a method lends itself well to closed-loop user-feedback, in which a user can adjust the value assigned to different features.

\subsection{Generating Explanations}
Given a near-optimal policy, here calculated with SARSOP \cite{kurniawati2008sarsop} (which we will consider consider optimal for the purposes of this work), a sequence of actions is executed following this policy, conditioned on observations, forming an apparent path. Given this path, a user may have questions as to why the path is optimal or an alternative path was not chosen. The user would then be prompted to provide an alternative path, represented as an open-loop action sequence in this work. Given the closed-loop optimal policy and the open-loop user policy, feature expectations can be calculated recursively using the Bellman expectation equation applied to the feature indicator function: $ \mu^\pi(b_t) = \phi(b_t,\pi(b_t)) + \gamma \, \mathbb{E}[\mu^\pi(b_{t+1})]$. These feature expectations can then be translated into a plain language explanation, contrasting the outcomes expected under the two policies. The primary focus of this work is the use of feature expectations to summarize the performance of user counterfactual policies, with the specifics of translation left to future work.

It is worth noting that while this method benefits from domains in which users may readily provide counterfactuals such as the SAR POMDP domain, it can be applied to any domain in which a user provides a counterfactual policy, whether open-loop or closed-loop. This method only requires that feature expectations can be defined and calculated for a counterfactual policy and the policy requiring explanation.

\section{Case Studies}

To demonstrate our approach, we formulate a SAR POMDP with a robot searching an $n x n$ grid for a partially observable stationary target while also visiting regions of interest with a limited battery capacity. The location of the target is unknown to the robot and user and is only inferred through noisy observations. There is a uniform initial belief $b$ over which cell the target occupies. The state space of this POMDP is made up of robot position $s_{robot}$, target position $s_{target}$, and remaining battery $s_{batt}$. Noisy observations of the target position are given if the robot is within one grid cell of the target. A perfect observation of the target location is provided if the robot is in the same cell as the target. Transitions are deterministic in position, with the robot moving in the direction indicated by any of the cardinal direction actions, while the battery deterministically decreases by $1$ with each action. The problem terminates if the target is found (the robot and target share the same cell) or if the difference between the remaining battery and the battery required to return to the starting location is less than 1. A reward $r_{target}$ is given for finding the target. Supplementary reward $r_{1:N}$ is given for visiting the $N$ locations of interest $l_{1:N}$.

For this SAR POMDP, consider the following general features. Let $l_{1:N}$ denote cells of interest (which would be specified by a user in a collaborative SAR task as discussed in Example \ref{example}), then the feature indicator function is defined as follows:
\begin{equation}
    \phi(s,a) = [x_1,... , x_N,x_t,x_b]
\end{equation} where
\begin{equation*}
    x_i = \begin{cases}
        1 \text{ if } s_{robot} = l_i\\
        0 \text{ o.w. }
    \end{cases} \forall i\in[1...N], \
        x_t = \begin{cases}
        1 \text{ if } s_{robot} = s_{target}\\
        0 \text{ o.w. }
    \end{cases},
\end{equation*}%
\begin{equation*}
    x_b = \begin{cases}
        1 \text{ if } (batt\_to\_go - s_{batt}) \leq 1\\
        0 \text{ o.w. }
    \end{cases},
\end{equation*}
and $batt\_to\_go$ is the battery required to return to the initial robot state from the current robot state. These features relate back to Example \ref{example}, representing user-specified objectives (cells of interest), a central objective (locating the target), and a constraint on the problem (preserving battery life such that the robot can return to base). Let the feature weighting be defined $\alpha = [r_1, ..., r_N, r_{target}, 0]$.

\subsection{Case Study 1: Observable and Unobservable Objectives}
The purpose of this example is to demonstrate contrastive explanations of paths in a context where there is one readily observable objective (a cell of interest) and one partially observable objective (the hidden target), the location of which is unknown initially.

\subsubsection{Model and Features}
For this example, there is one cell of interest $l_1=[1,5]$ with reward $r_1=3.0$ and a partially observable target located at $s_{target}=[5,5]$ with reward $r_{target}=500.0$. The available battery is $s_{batt}=25.0$.

\begin{figure*}
    \centering
    \begin{subfigure}{0.24\linewidth}
        \centering
        \includegraphics[width=0.75\linewidth]{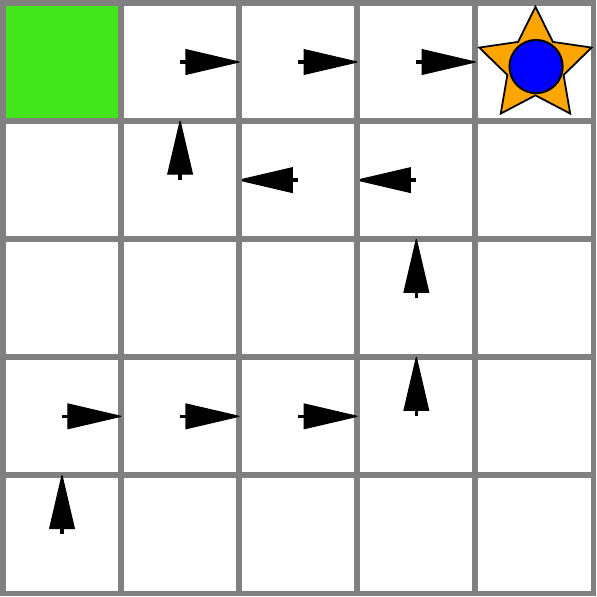}
        \caption{Optimal Policy}
        \label{fig:demo1_sar}
    \end{subfigure}%
    \begin{subfigure}{0.24\linewidth}
        \centering
        \includegraphics[width=0.75\linewidth]{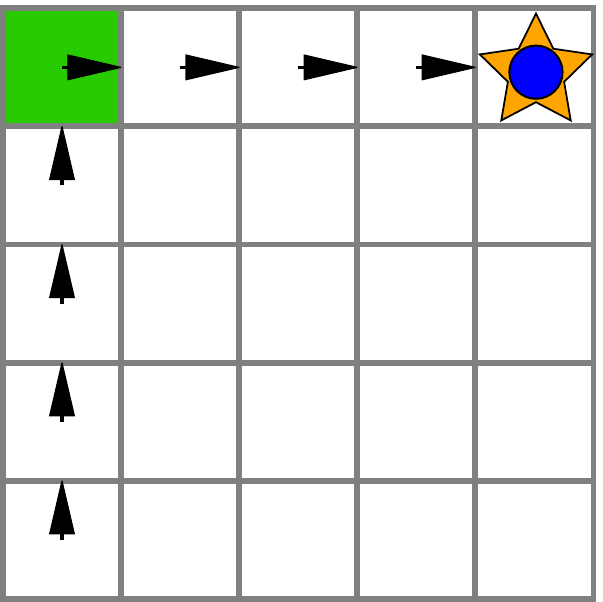}
        \caption{User Policy}
        \label{fig:demo1_hu}
    \end{subfigure}
    \begin{subfigure}{0.24\linewidth}
        \centering
        \includegraphics[width=0.75\linewidth]{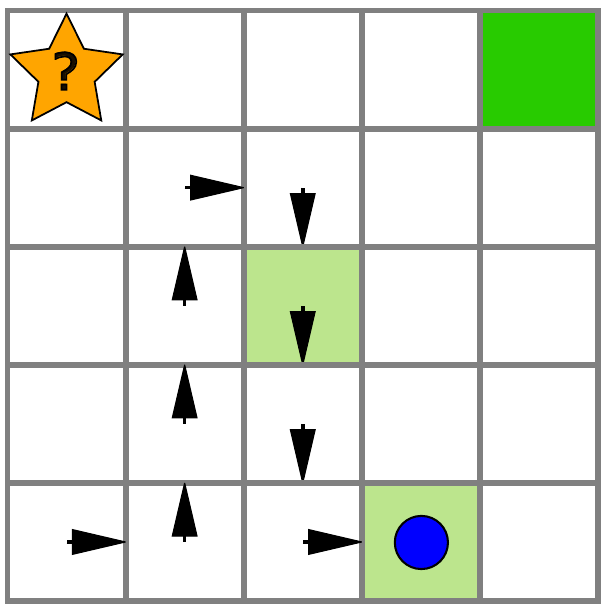}
        \caption{Optimal Policy}
        \label{fig:demo2_sar}
    \end{subfigure}%
    \begin{subfigure}{0.24\linewidth}
        \centering
        \includegraphics[width=0.75\linewidth]{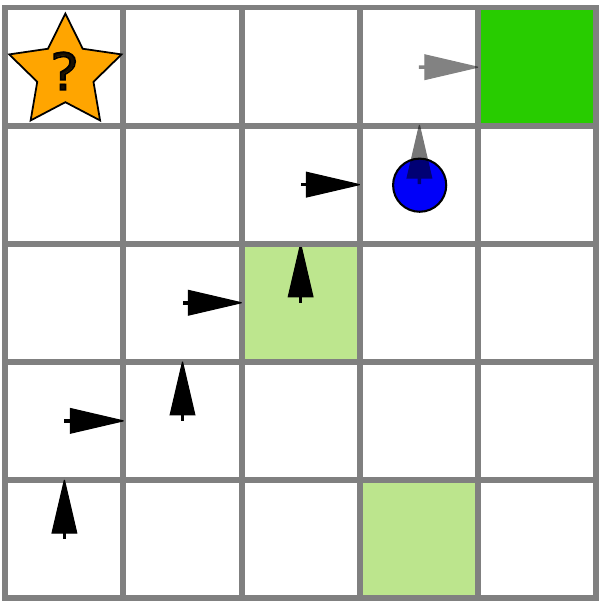}
        \caption{User Policy}
        \label{fig:demo2_hu}
    \end{subfigure}
    \label{fig:demos}
    \caption{Case studies. (a), (b) An example in which the readily observable objective and the more valuable, partially observable, objective do not align. Note the target location (orange star) is unknown initially and only discovered by the robot (blue circle) after the optimal policy is executed. (c), (d) An example in which constraints restrict the feasibility of a proposed user policy. The \textit{black} arrows represent the executed actions while the \textit{gray} arrows represent the remaining actions of the user counterfactual path that were not executed due to the agent reaching a terminal battery state.}
\end{figure*}

\subsubsection{Contrasting Path Outcomes}
In this domain, the optimal policy executes the path shown in Fig. \eqref{fig:demo1_sar} and finds the target, which is now visible to the user. With this hindsight knowledge of the location of the target, the user may wonder why the optimal policy did not simply go immediately up to collect the observable reward, and then to the target, as in Fig. \eqref{fig:demo1_hu}. In fact, for this particular simulation, the user policy (with hindsight knowledge of the target location) achieves greater discounted reward ($r^{\pi_{hu}} = 334.154$) than the optimal policy ($r^{\pi^*} = 270.180$). This further underscores the need for explanation. The feature expectations from the open-loop policy based on the suggested path $\pi_{hu}$ and the optimal SARSOP policy $\pi^*$ are shown in Table \eqref{tab:demo1}.

\begin{table}[h]
\centering
\caption{Feature expectations for optimal and open-loop user policies for Case Study 1.}
\label{tab:demo1}
\begin{tabular}{|c | c c c |} 
    \hline
    & $l_1$ & $target$ & $battery$  \\
    \hline
    $\mu^{\pi^*}$ & 0.036 & $\mathbf{0.731}$ & 0.0\\
    \hline
    $\mu^{\pi_{hu}}$ & $\mathbf{0.684}$ & 0.296 & 0.0 \\
    \hline
\end{tabular}
\end{table}

While in this case, where the target is in cell $[5,5]$, both policies find the target and the user policy achieves greater discounted reward than the optimal policy, the user policy does not account for the uncertainty in target location. In expectation over all possible target states (i.e. the initial belief), the closed-loop optimal policy outperforms the open-loop user policy in terms of the expected frequency of locating the target. This results in the optimal policy having a higher value than the user policy, as locating the target has a much higher weight. The open-loop user policy does outperform the optimal policy in terms of the frequency with which it reaches the cell of interest, but this feature has a much lower weight.

Because the target could be in any one of the cells and the problem terminates when the robot and target share a cell, there is some probability of terminating in every cell from the initial belief. This is likely why frequencies less than $1$ are observed.

In order to produce an explanation, these feature occupancies must be translated into plain language, as in \cite{khan2009minimal}. The translation approach itself is left to future work. A plausible explanation based on the feature expectations could be:

\begin{quote}
    \textit{"Over all possible target locations, \textbf{the optimal policy finds the target about twice as often} as the user policy. The optimal policy will visit the cell of interest almost never. Since \textbf{the target has a much higher weighting than the cell of interest}, the optimal policy will outperform the user policy."}
\end{quote}

\subsection{Case Study 2: Resource Constraints}
The purpose of this example is to demonstrate the effectiveness of contrastive feature expectation explanations as they relate to constraints on the problem, in this case the finite battery available in the SAR POMDP.

\subsubsection{Model and Features}
For this example, there are three cells of interest $l_1=[5,5]$, $l_2=[4,1]$, and $l_3=[3,3]$ with reward $r_1=3.0$, $r_2=1.0$, and $r_3=1.0$, respectively. A partially observable target is located at $s_{target}=[1,5]$ with reward $r_{target}=100.0$. The available battery is $s_{batt}=12.0$.

\subsubsection{Contrasting Path Outcomes}
Given the path generated by the optimal policy (Fig. \eqref{fig:demo2_sar}), the user may wonder why the path did not reach the higher-reward cell of interest in the upper right ($l_1$) and propose a path to that cell. However, the battery constraint does not allow for that cell to be reached and the shortened, feasible path shown Fig. \eqref{fig:demo2_hu} is used as the basis for comparing outcomes. The corresponding feature expectations are shown in Table \eqref{tab:demo2}.

\begin{table}[h]
\centering
\caption{Feature expectations for optimal and open-loop user policies for Case Study 2.}
\label{tab:demo2}
\begin{tabular}{|c | c c c c c|} 
    \hline
    & $l_1$ & $l_2$ & $l_3$ & $target$ & $battery$  \\
    \hline
    $\mu^{\pi^*}$ & 0.0 & $\mathbf{0.202}$ & 0.354 & $\mathbf{0.550}$ & $\mathbf{0.346}$\\
    \hline
    $\mu^{\pi_{hu}}$ & 0.0 & 0.0 & $\mathbf{0.684}$ & 0.241 & 0.559\\
    \hline
\end{tabular}
\end{table}

From these feature expectations, it is apparent that neither policy successfully reaches the higher-reward cell ($l_1$) and that the optimal policy is about twice as likely to locate the target when compared to the open-loop user policy. Likewise, the optimal policy avoids the battery terminal criteria more often. The optimal policy will visit the other cells of interest ($l_2,l_3$) in aggregate slightly less often than open-loop user path.

With these feature expectations and domain-knowledge about the limited battery, an explanation of the following form could be provided:

\begin{quote}
    \textit{"The \textbf{battery constraint} makes it \textbf{impossible for either policy to reach} $l_1$. Over all possible locations of the target, \textbf{the optimal policy will find the target more often} leading to a higher reward (since the target is valued higher than any location of interest)."}
\end{quote}

\section{Conclusion and Future Direction}
In this work, we present an approach to explaining paths generated by optimal solutions to POMDP search and rescue problems. This initial approach takes a counterfactual user path as the basis for an open-loop policy which is contrasted against an optimal policy through the use of feature expectations.

While this work presents one means of providing contrastive explanations of optimal POMDP solutions, there are shortcomings to this approach. One substantial assumption is that the user maintains an open-loop policy that does not change with new information. However, accounting for policy changes due to new information likely better captures user reasoning. In particular, accounting for the influence of observations on user's reasoning will likely make for more effective POMDP explanations which better capture user behavior. Ideally, such methods could account for this closed-loop reasoning while still requiring limited user input.

Additionally, providing proactive explanations of executed paths which can be provided automatically in anticipation of user confusion will be valuable in reducing user workload. This would also reduce dependence on a domain-specific means of user feedback.


\bibliographystyle{ACM-Reference-Format}
\bibliography{sample-base}

\end{document}